\documentclass[10pt, conference, compsocconf]{IEEEtran}
\usepackage{cite}

\ifCLASSINFOpdf
  \usepackage[pdftex]{graphicx}
  \DeclareGraphicsExtensions{.pdf,.jpeg,.png,.jpg}
\else
\fi
\usepackage[font=footnotesize,caption=false]{subfig}
\begin{document}
%
\title{A Framework for Video-Driven Crowd Synthesis\\
{\small (Supplementary material available at https://faisalqureshi.github.io/research-projects/crowds/crowds.html)}
}


\author{\IEEEauthorblockN{Jordan Stadler}
\IEEEauthorblockA{Inscape Corp.\\
Holland Landing ON Canada}
\and
\IEEEauthorblockN{Faisal Z. Qureshi}
\IEEEauthorblockA{Faculty of Science\\
University of Ontario Institute of Technology\\
Oshawa ON Canada\\
Email: faisal.qureshi@uoit.ca}
}


%


\maketitle

\begin{abstract}
We present a framework for video-driven crowd synthesis.  Motion
vectors extracted from input crowd video are processed to compute
global motion paths.  These paths encode the dominant motions observed
in the input video.  These paths are then fed into a behavior-based
crowd simulation framework, which is responsible for synthesizing
crowd animations that respect the motion patterns observed in the
video.  Our system synthesizes 3D virtual crowds by animating virtual
humans along the trajectories returned by the crowd simulation
framework.  We also propose a new metric for comparing the ``visual
similarity'' between the synthesized crowd and exemplar crowd.  We
demonstrate the proposed approach on crowd videos collected under
different settings.
\end{abstract}


%
\IEEEpeerreviewmaketitle

\section{Introduction}

Raynold's seminal 1987 paper on {\it boids} showcased that group
behaviors emerge due to the interaction of relatively simple,
spatially local rules~\cite{reynolds1987flocks}.  Since then there has
been much work on crowd synthesis.  The focus has been on methods for
generating crowds exhibiting highly realistic and believable motions.
Crowd synthesis, to a large extent, remains the purview of computer
animators, who painstakingly fiddle with numerous parameters in order
to achieve believable crowd motions.  Many animation tools exist for
generating high-quality crowds for computer entertainment industries.
MASSIVE~\cite{massive14}, Golaem Crowd~\cite{golaem14},
Miarmy~\cite{miarmy14}, and Maya~\cite{maya14}, for example, are
popular crowd animation tools.  All of these tools have steep learning
curves and these require a lot of manual tweaking to animate crowds
having the desired motion/appearance characteristics.  Exemplar-based
crowd synthesis appears a promising direction of future
research~\cite{li2012cloning}.  Here, crowd synthesis parameters are
learned by observing ``real'' crowds.  These parameters can
subsequently be used to synthesize crowds in previously unseen
settings, i.e., different viewpoints, new environmental settings, etc.

\begin{figure}
        \centerline{ \includegraphics[width=0.33\linewidth,
            height=2cm]{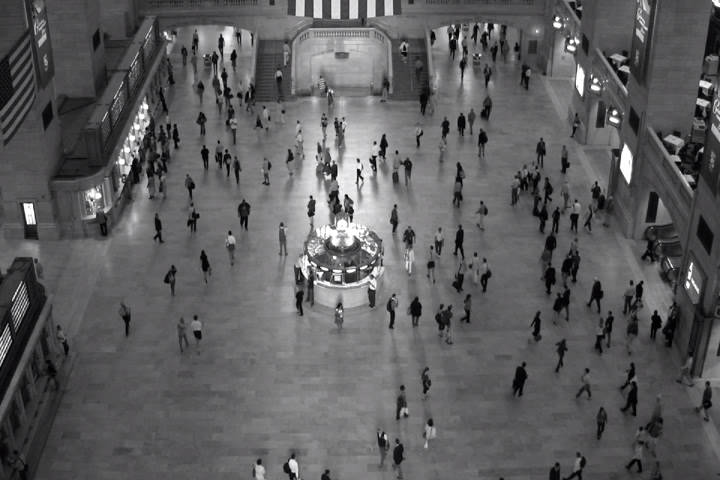}\hfill
          \includegraphics[width=0.33\linewidth,
            height=2cm]{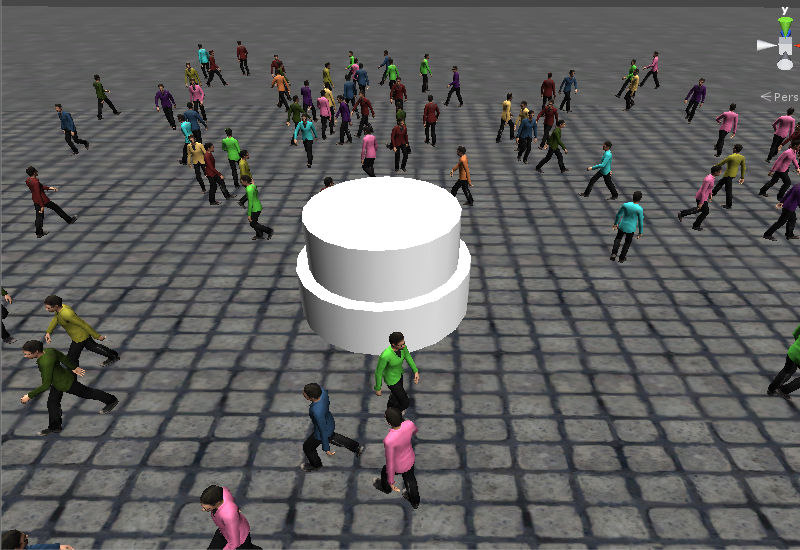}\hfill%
        \includegraphics[width=0.33\linewidth, height=2cm]{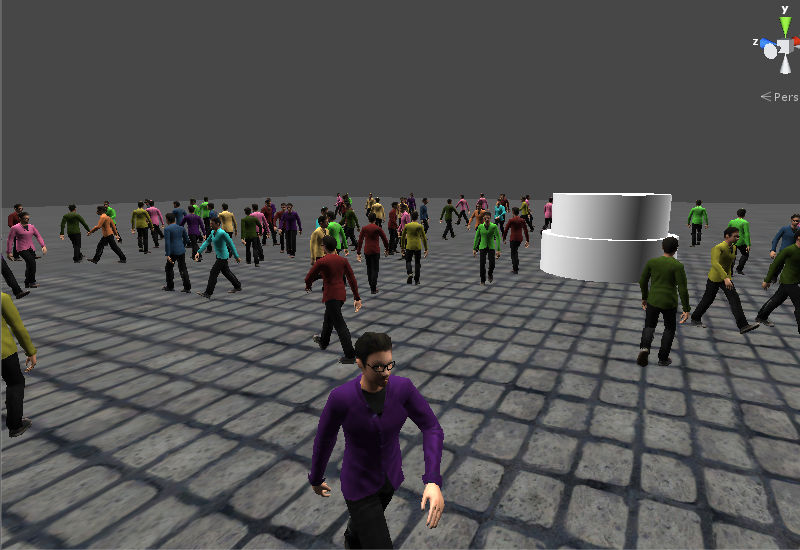}}

        \centerline{ \includegraphics[width=0.33\linewidth,
            height=2cm]{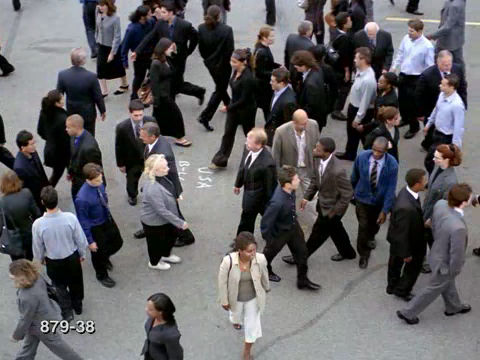}\hfill
          \includegraphics[width=0.33\linewidth,
            height=2cm]{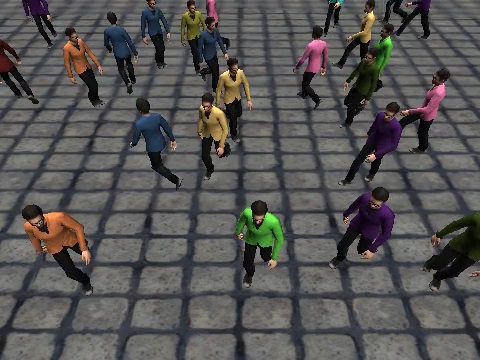}\hfill%
        \includegraphics[width=0.33\linewidth,
          height=2cm]{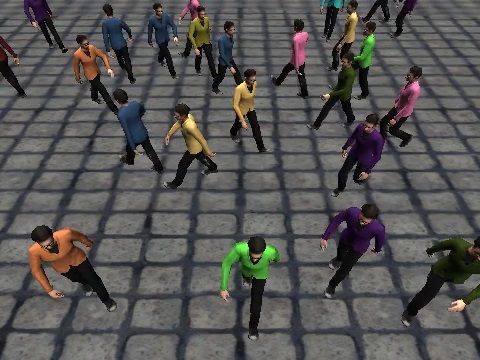}}

        \centerline{ \includegraphics[width=0.33\linewidth,
            height=2cm]{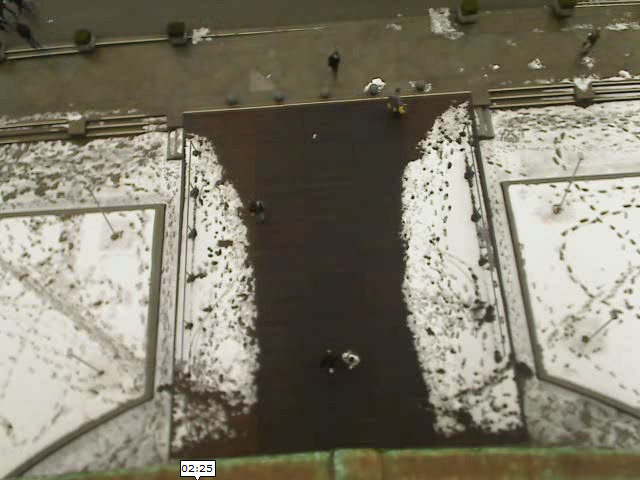}\hfill
          \includegraphics[width=0.33\linewidth,
            height=2cm]{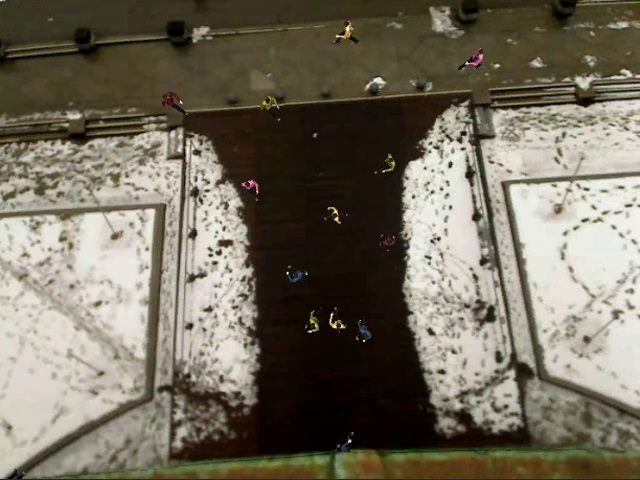}\hfill%
        \includegraphics[width=0.33\linewidth,
          height=2cm]{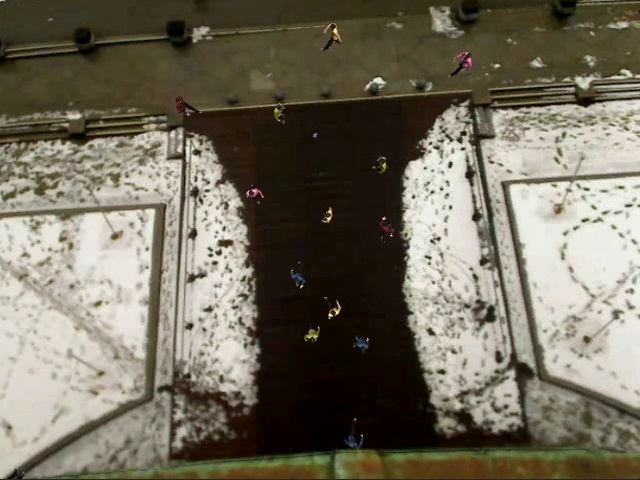}}

        \caption{ Virtual crowd synthesized by analyzing an exemplar
          videos. Top to bottom: Grand central, Campus and UCF Walking
          dataset.  (Left) A frame from exemplar video. (Middle and
          right) Frames from synthesized crowds.}

        \label{fig:multiview}
\end{figure}

\begin{figure}
        \centerline{
        \includegraphics[width=0.16\linewidth]{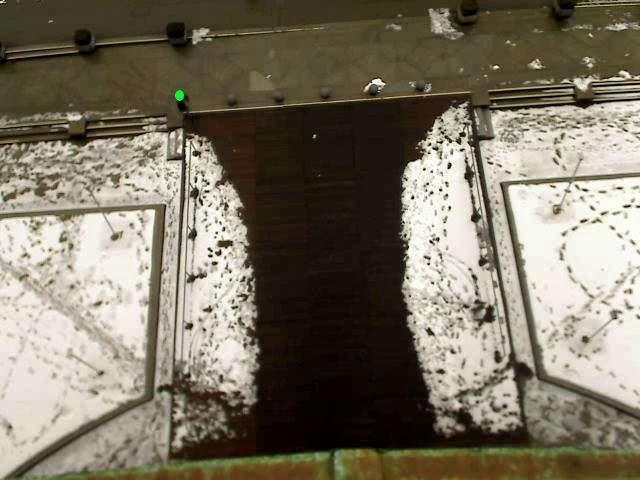}
        \includegraphics[width=0.16\linewidth]{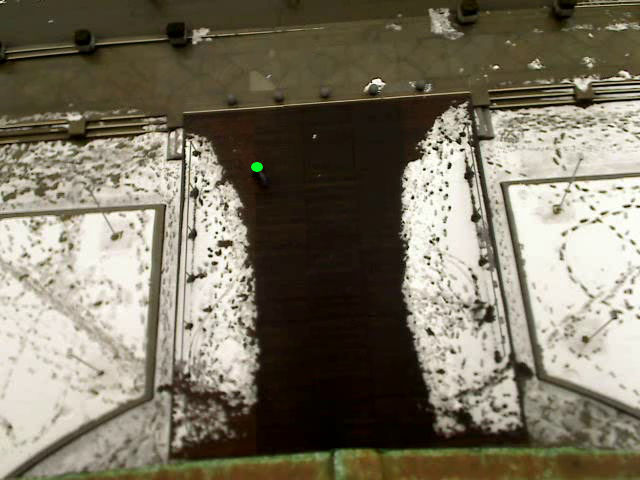}
        \includegraphics[width=0.16\linewidth]{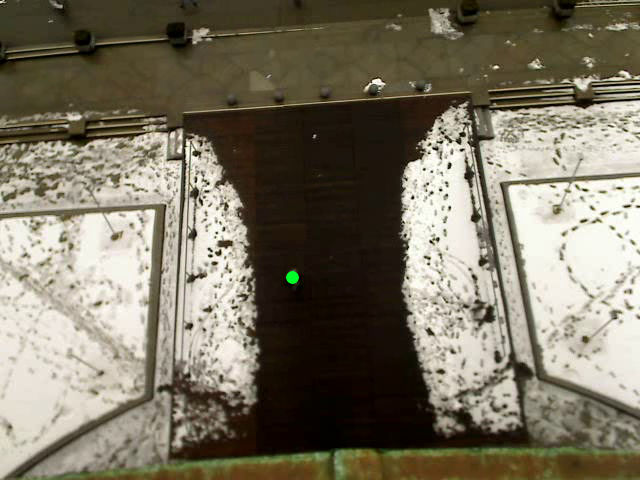}
        \includegraphics[width=0.16\linewidth]{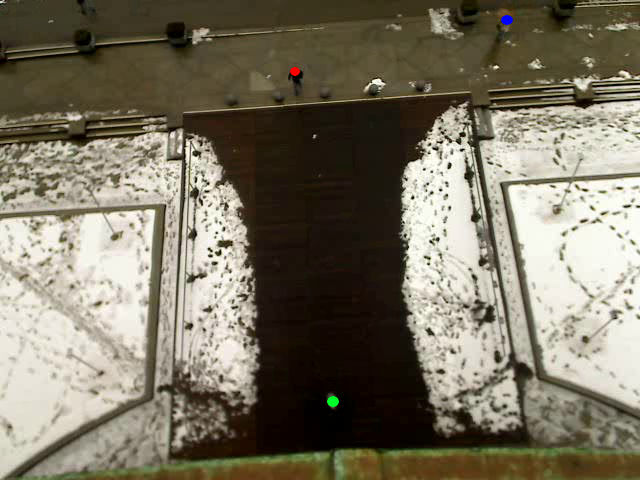}
        \includegraphics[width=0.16\linewidth]{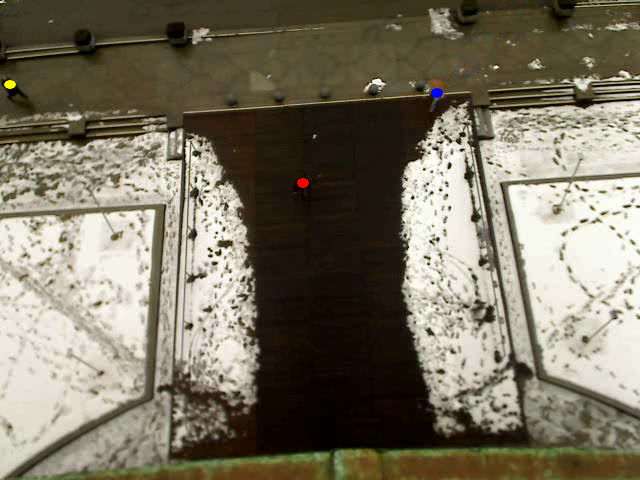}
        \includegraphics[width=0.16\linewidth]{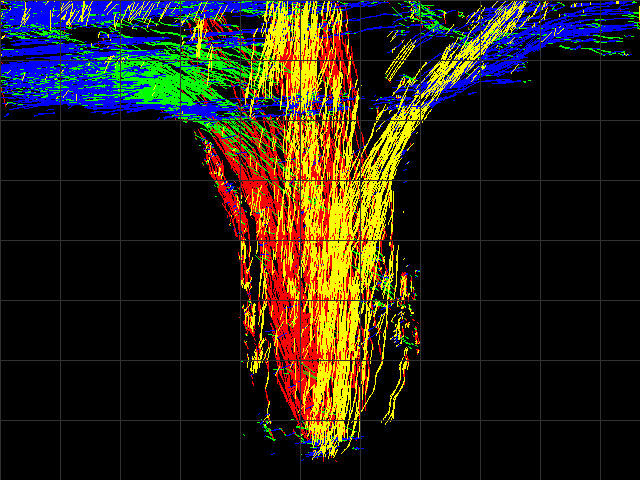}}
\centerline{
        \includegraphics[width=0.16\linewidth]{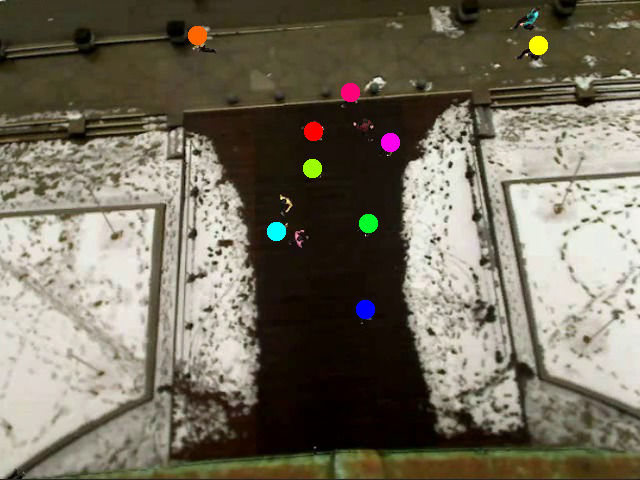}
        \includegraphics[width=0.16\linewidth]{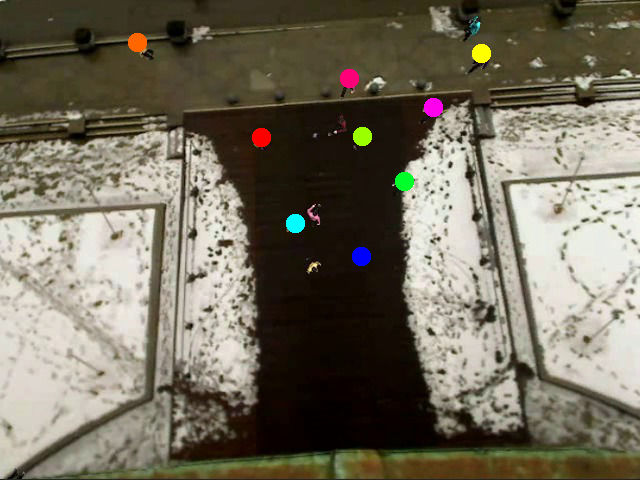}
        \includegraphics[width=0.16\linewidth]{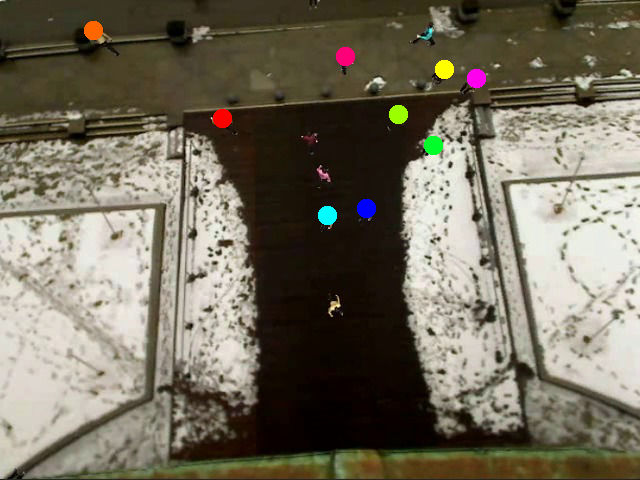}
        \includegraphics[width=0.16\linewidth]{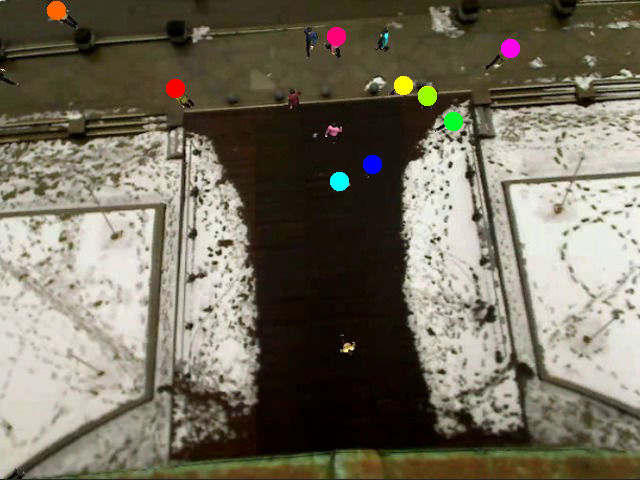}
        \includegraphics[width=0.16\linewidth]{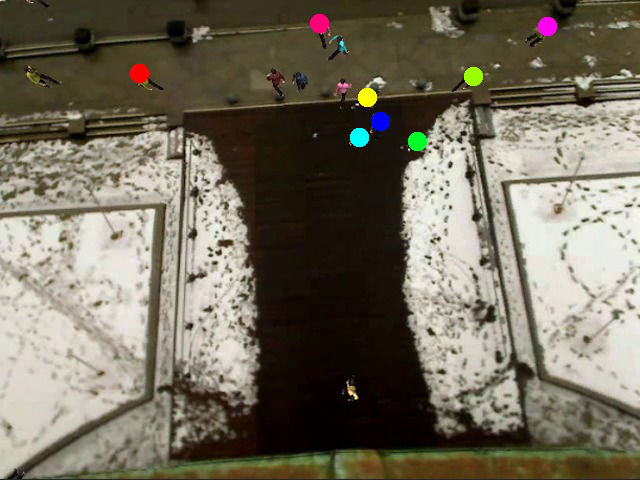}
        \includegraphics[width=0.16\linewidth]{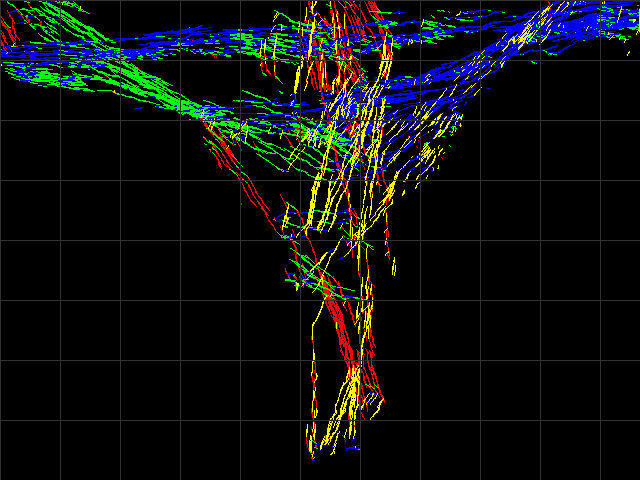}}
        \caption{ Top row: frames from the exemplar video and
          (right) extracted motion vectors.  Bottom row: stills
          from the synthesized crowd and
          (right) extracted motion vectors.}
        \label{fig:compare}
\end{figure}

Within this context, this paper develops a framework for crowd
synthesis via analysis (Figure~\ref{fig:multiview}).  Videos
exhibiting crowds are analyzed to extract high-level motion patterns.
These motion patterns are then combined with (spatially) local
behavior rules, such as collision avoidance, path following, velocity
matching, etc., to synthesize crowds.  Specifically, we use Reciprocal
Collision Avoidance for Real-Time Multi-Agent Simulation (RVO2) to
synthesize crowd animations given the constraints extracted from
exemplar videos~\cite{van2008reciprocal}.  RVO2 provides us with
trajectories for individual agents and we use motion graphs to animate
3D virtual humans moving along these
trajectories~\cite{kovar2002motion}.  Figure~\ref{fig:compare} shows
frames from videos of real and synthesized crowds and the motion
information extracted from these videos to compare the crowds seen in
these videos.

Crowd analysis is an active field of research within the computer
vision community.  Many species, including humans, exhibit coordinated
group behavior: schools of fish, flocks of birds, herds and packs of
animals, and human crowds~\cite{sumpter2006principles}.  Some argue
that such group behaviors are important for
survival~\cite{bode2010perceived}.  Humans also have a great facility
for perceiving group behavior~\cite{thornton2004incidental}.  There is
mounting evidence from the psychology literature that humans are able
to perceive group behavior without decoding the individual motions.
This work explicitly assumes that individual motions (i.e.,
trajectories of individuals seen in the exemplar videos) are {\it not}
available.

We also introduce a new metric for comparing the original crowd with
a synthesized crowd.  In order to compare the synthesized crowd with the
original crowd, we render the synthesized crowd from a viewpoint
similar to the one used to record the video of the real crowd.  Motion
parameters are extracted from the rendered footage and compared with
those extracted from the real footage.  Preliminary results seem to
suggest that this metric is able to rank crowd pairs according to
their motion similarities.  More work is needed to further study this
aspect of this work.  The ability to compute the similarity between
two crowds is needed when setting up a feedback loop that
iteratively refines the synthesized crowds to better match real
crowds.

\section{Related Work}

{\bf Crowd analysis} generally falls into one of three categories:
density estimation, pedestrian tracking, and behavior detection \&
recognition.  Crowd density, although a useful metric for determining
how occupied an area is, is not used in this system presently.
Pedestrian tracking~\cite{eshel2010tracking} is useful for
observing the movement of individial members in a crowd.  Behavioral
understanding, such as crowd event detection, can be used to assist
with surveillance systems by alerting of suspicious or dangerous
events.

Many approaches to pedestrian tracking suffer from issues related to
occlusions.  Eshel et al.~\cite{eshel2010tracking} propose a method of
minimizing these problems through multiple overhead cameras.  Head
tracking is performed across multiple videos to produce the pedestrian
tracks through the scene.  Although the results are promising, this
limits the crowd analysis to crowds that have been observed by these
multi-camera setups.  This paper strives to work with single videos of
crowds that are as unconstrained as possible.  For the purposes of
this work, we refer the kind reader to
~\cite{hu2008learning,ozturk2010detecting} that discuss estimation
global motion fields from crowd videos.

Motion tile/patch based crowd synthesis
approaches~\cite{yersin2009crowd,shum2008interaction,kim2012tiling,lee2006motion}
have received popularity in recent years for their scalability.  These
approaches to crowd synthesis are great for large-scale dynamic crowds
however they are not ideal for synthetic crowd reproduction.  Although
these crowds are interesting to the viewer, they are not well suited
for reproducing a crowd in that their focus is simply optimizing the
actions that are occurring for and among agents.  Velocity
fields~\cite{patil2011directing,wang2008data,chenney2004flow} offer
quality agent navigation results with the benefit of offering
navigation from any position in the environment.  The work of Patil et
al.~\cite{patil2011directing} offers a solution to directing crowd
simulations using velocity fields.  Similarly Wang et
al.~\cite{wang2008data} perform direct crowd simulation with velocity
fields generated from videos.

Flagg et al.~\cite{flagg2013video} propose a video-based crowd
synthesis technique utilizing crowd analysis for the purpose of
generating crowd videos with realistic behavior and appearance.  Their
approach generates pedestrian sprites by segmenting pedestrians from
input video and crowd tubes are used to avoid collisions and ensure
accurate agent navigation.  Their approach produces promising results
but it is constrained to the field of view of the original video.
Butenuth et al.~\cite{butenuth2011integrating} also produce a
simulation restricted to 2D, with discs representing agents.  Their
approach focuses on more dense crowds.

3D
approaches~\cite{li2012cloning,lee2007group,lerner2009fitting,sun2011data}
to synthesis via analysis offer flexibility in that they can be
heavily manipulated and customized by the end-user.  However, previous
approaches rely on very constrained input video.  The work of Lee et
al.~\cite{lee2007group} relies on a top-down facing camera to observe
the crowd and extract trajectories.  Lerner et
al.~\cite{lerner2009fitting} make use of this technique but also rely
on user input to annotate extracted trajectories for the purpose of
agent behavior detection and recognition.  Similarly having a reliance
on motion capture \cite{li2012cloning,sun2011data} data to feed a
simulation can be restrictive as input.  Ideally a system would be
capable of accepting an unconstrained video of a crowd and being able
to reproduce it, which is the focus of this paper.

Methods for crowd entropy, a measure of a crowd's disorder, can be
useful as a metric for observing and identifying crowd activities.
Various methods for calculating crowd entropy have been proposed and
used for different purposes.  Guy et al.~\cite{guy2012statistical}
propose a method for computing an entropy score for a given crowd
navigating through a scene.  Their method is used to evaluate steering
methods and requires real-world data to compare the motions of an
individual agent.  The proposed system is more interested in comparing
the output video of the pipeline versus the input crowd video.
Ihaddadene et al.~\cite{ihaddadene2008real} perform real-time crowd
motion analysis in which they maintain an entropy value to watch for
specific variations.  Their method is not used for the evaluation of
crowds but by observing the entropy they can estimate sudden changes
and abnormalities in the crowd's motion.

It is not uncommon for crowd simulations to be evaluated with a visual
comparison performed by study
groups~\cite{Lee:2007:GBV:1272690.1272706, karamouzas2012simulating}.
Performing study group visual comparisons is lengthy and does not
leave automation as a possibility, which is a goal for this system.

\section{Crowd Analysis}

We have experimented with three schemes for extracting motion
information from crowd videos: a) dense optical flow estimation
technique proposed by Farneback~\cite{farneback2003two}, b) sparse
optical flow estimation~\cite{ozturk2010detecting}, and c) SIFT based
tracking.  We found that the sparse optical flow estimation technique
serves our purpose well.  It is also the most efficient of the three.
This method is also suitable for dense crowds, since it doesn't assume
that individual members are tracked.

  \begin{figure}
    \centerline{
      \includegraphics[width=2.5in]{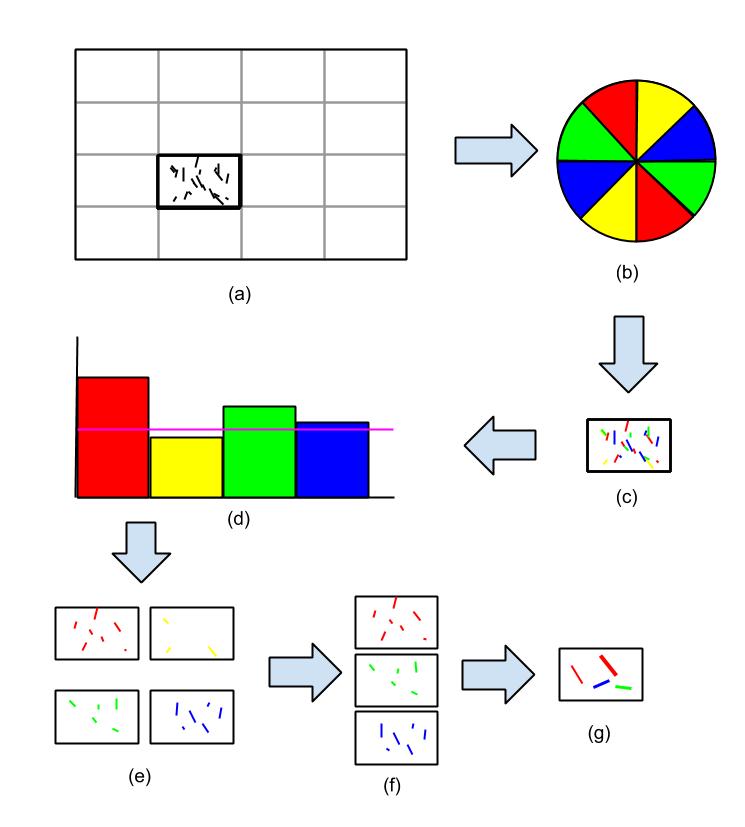}}
  \caption{ Stages of clustering.
    (a) scene is divided into cells.  (b) motion vectors per cell are
    assigned to one of 4 orientation clusters (demonstrated with one
    of 4 colors). (c) a single cell and it's motion vectors.  (d)
    orientation clusters per cell are used to generated a histogram of
    orientations. Weak orientations are discarded.  (e) a single cell
    and its per orientation vectors.  (f) showing orientations for a
    single cell that are kept.  These enter self-tuning spectral
    clustering.  (g) The resulting locally dominant directions.}
    \label{fig:order}
  \end{figure}

Given a sequence of (key)frames $I_1,I_2,I_3,\cdots,I_n$, inter-frame
motion extraction returns a set of motion vectors $(x,y,u,v,t)$, where
$t$ refers to the frame id and $t \in [1,n-1]$.  Motion vectors are
stored as $(x,y,\theta,l,t)$, where $\theta = \arctan \left(
\frac{y}{x} \right)$ and $l = \sqrt{x^2+y^2}$.  This representation
facilitates orientation-based grouping of motion vectors.

The goal is to combine these motion vectors to construct dominant
paths.  This is accomplished, following the work of Ozturk et
al.~\cite{ozturk2010detecting}, through binning, pruning and
clustering steps.  The image space is first divided into cells
(Figure~\ref{fig:order})---cell extents are defined in pixel
locations.  Motion vectors belonging to the same cell are aggregated
into 8-bin orientation histograms $H^{(i,j)}_\theta$.  $(i,j)$ here
refer to the location of the spatial bin---in the example shown in
Figure~\ref{fig:order}, $i \in [1,40]$ and $j \in [1,40]$---and
$H^{(i,j)}_\theta(k)$ refers to the $k^{\mathrm{th}}$ bin of this
histogram, where $k \in [1,8]$.  After this step each cell is
represented by an 8-bin orientation histogram.  Vectors belonging to
diagonally opposite bins in the orientation wheel are shown in the
same color.  Following the work of Ozturk et
al.~\cite{ozturk2010detecting}, 8-bin orientation histograms yield
acceptable results.  However, it is straightforward to change the
number of bins.  Aggregating nearby motion vectors into orientation
histograms has a desirable side-effect.  It allows us to discard
motion vectors that fall in orientation bins with little support,
i.e., if the number of motion vectors in a particular orientation bin
is less than a threshold, it can safely be ignored for that direction
(for that spatial location) in subsequent processing).

\subsubsection{Spectral Clustering}

\begin{figure}
    \centerline{
      \subfloat[]{\includegraphics[width=.5\linewidth]{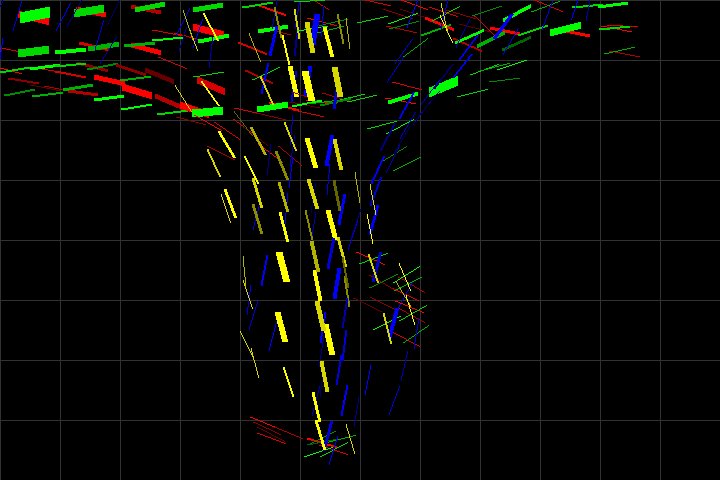}}
      \subfloat[]{\includegraphics[width=.5\linewidth]{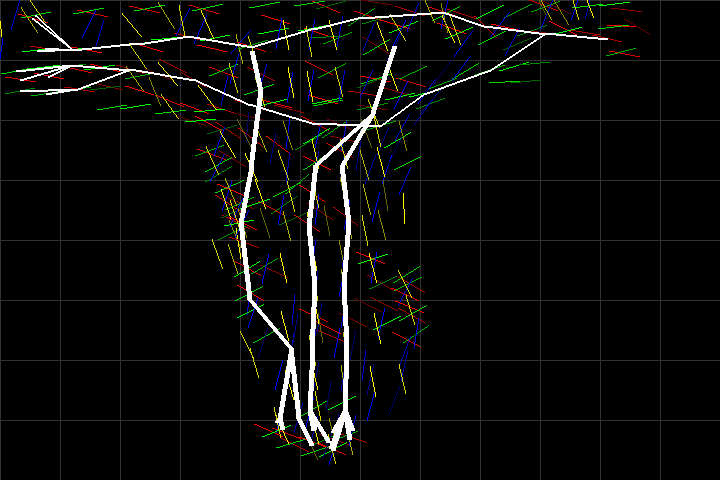}}}
  \caption{ (a) Spectral clustering returns spatially local dominant
    directions.  The color of the vector represents its orientation;
    whereas, its thickness indicates the number of motion vectors
    that belong to this cluster. (b) Globally dominant paths
    constructed from spatially local dominant directions using the
    strategy described in Ozturk et al.~\cite{ozturk2010detecting}}
    \label{fig:domvecs-paths}
  \end{figure}

Motion vectors within orientation histogram bins that survive pruning
are then clustered to compute spatially local dominant directions (See
Figure~\ref{fig:order}).  The Self-Tuning Spectral Clustering scheme
proposed by Zelnik et al.~\cite{zelnik2004self} is used for this. The
affinity matrix is computed as follows:
\[
A(m,n) = \exp \left( - \frac{ \left| \mathbf{p}_m - \mathbf{p}_n
    \right|^2 }{\sigma_m \sigma_n} \right),
\]  
where $\mathbf{p}_m$ and $\mathbf{p}_n$ represent spatial locations
$(x,y)$ of the $m^\mathrm{th}$ and $n^\mathrm{th}$ motion vectors in
bin $H^{(i,j)}_\theta(k)$. $\sigma_m$ and $\sigma_n$ represent scale
values.
Specifically, $\sigma_m$ is the Euclidean distance between
$\mathbf{p}_m$ vector and its $k^{\mathrm{th}}$-nearest neighbor (in
the same orientation histogram bin), and $\sigma_n$ is the Euclidean
distance between $\mathbf{p}_n$ vector and its
$k^{\mathrm{th}}$-nearest neighbor.  Specifically we use the $7^{\mathrm{th}}$-nearest
neighbor is used when computing these values.
The details of this algorithm are found in Zelnik et
al.~\cite{zelnik2004self}.  Clustering yields spatially local dominant
directions $(x,y,\theta,w)$, where $(x,y)$ represent the position,
$\theta$ denotes the orientation, and $w \in [0,1]$ indicates the
weight (or support) for that direction~(Figure~\ref{fig:domvecs-paths}(a)).

\subsection{Path Generation}

The final step is to combine these locally dominant vectors into
global paths using the approach described in Ozturk et
al.~\cite{ozturk2010detecting}.  Given a (dominant) direction vector,
search in its neighboring cells to find vectors having similar
orientations and group the two vectors to grow the path.  If the
neighboring cells do not contain a vector with similar orientation,
then consider vectors in other orientations.  In practice the cells
are swept from left-to-right and from top-to-bottom to grow dominant
direction vectors into global paths. Different sweeping methods can be
used (such as opposite directions) and smaller temporal chuncks can be
processed and combined if more paths are
desired. Figure~\ref{fig:domvecs-paths}(b) illustrates global paths
generated from locally dominant directions returned by the spectral
clustering procedure.

\section{Crowd Synthesis}
\label{sec:crowd-syn}

For crowd synthesis, RVO2 is used to provide a behavior-based agent
simulation system to simulate the movement of agents on a 2D plane.
3D virtual humans are animated along the trajectories returned by the
RVO2 simulator.  RVO2 implements a behavior-based multi-agent
simulation framework.  Every agent is treated as an autonomous entity,
complete with perception, decision-making, and action routines.
Perception routines enable an agent to ``observe'' its environment,
identify other agents, obstacles, and other items of interest.
Decision-making takes into account the current state of the agent, its
surroundings, and its goals.  It then updates the current position and
velocity of the agent.  RVO2 implements start-of-the-art collision
avoidance and goal arrival behaviors.

Each simulated agent has a unique goal stack which contains the
information needed to navigate the scene successfully.  Goal stacks
currently contain nodes representing locations along the path that
an agents must take.  The stack updates as each goal is met.  An empty
goal stack suggests that the agent has reached the end of its path.
At this point the agent can be safely removed from the simulation.
Although the goal stack is used in this system purely for navigation,
it can also be used to support {\it layered} behavior.  Similarly
subgoals can be pushed onto the stack to ensure they are completed
first.  Subgoals are currently used in the system to achieve smoother
agent trajectories.  Curve interpolation is used as a path smoothing
method for agent navigation by providing interpolated subgoals.  By
navigating the subgoals, the agent is able to advance toward their
goals with a more natural approach.

\begin{figure}
  \centerline{
  \subfloat[]{\includegraphics[width=.33\linewidth]{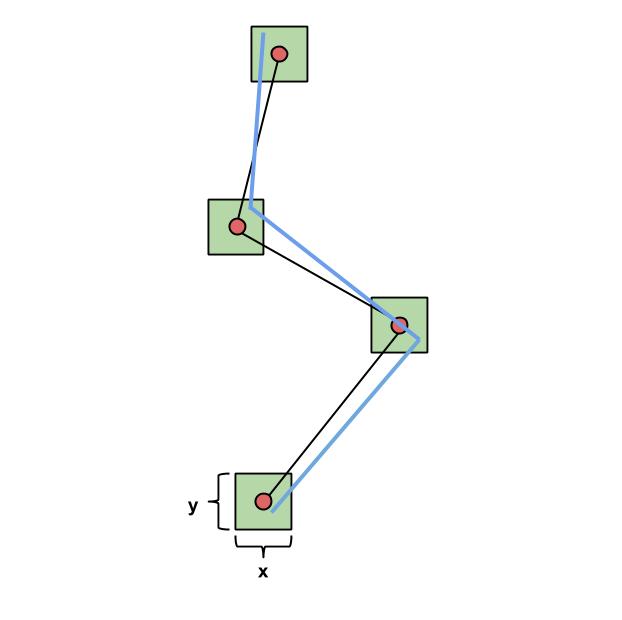}} \hfill
  \subfloat[]{\includegraphics[width=.33\linewidth]{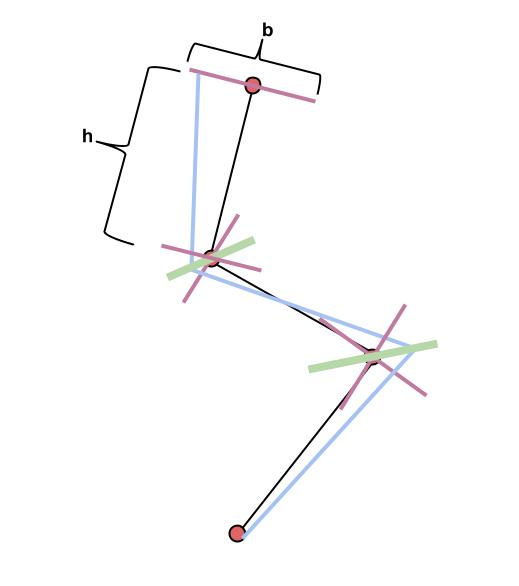}}
  \subfloat[]{\includegraphics[width=.33\linewidth]{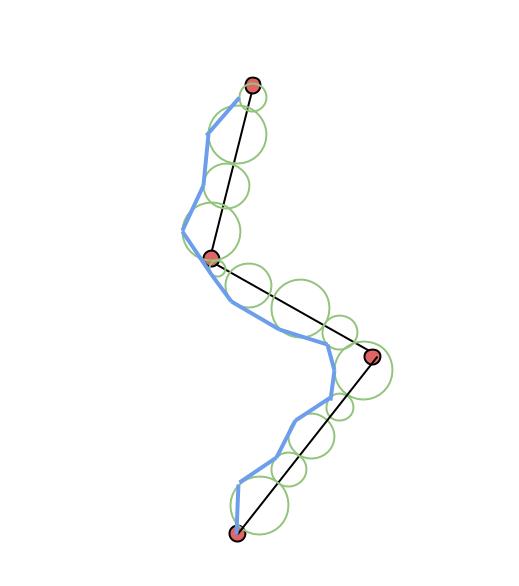}}
  }
  \centerline{
  \subfloat[]{\includegraphics[width=.33\linewidth]{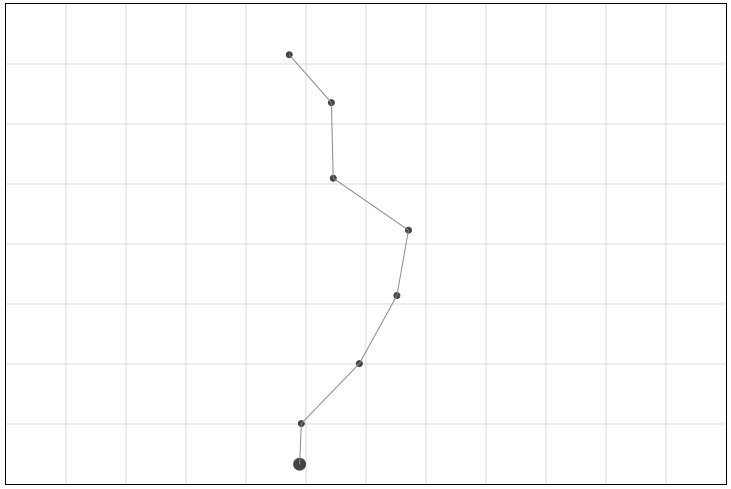}} \hfill
  \subfloat[]{\includegraphics[width=.33\linewidth]{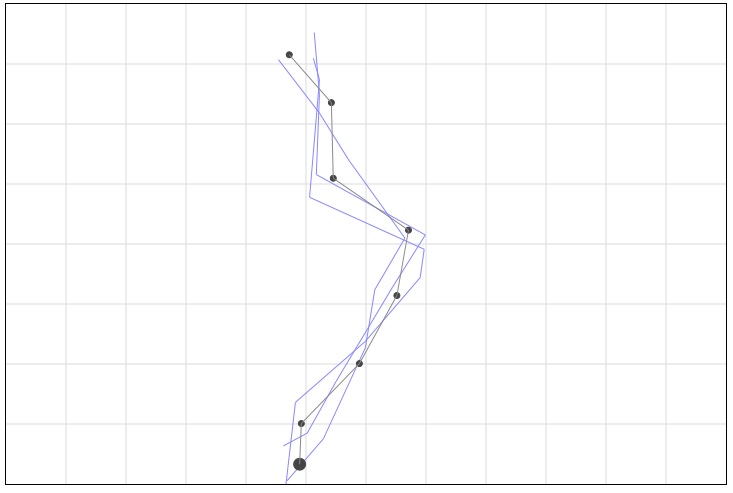}} \hfill
  \subfloat[]{\includegraphics[width=.33\linewidth]{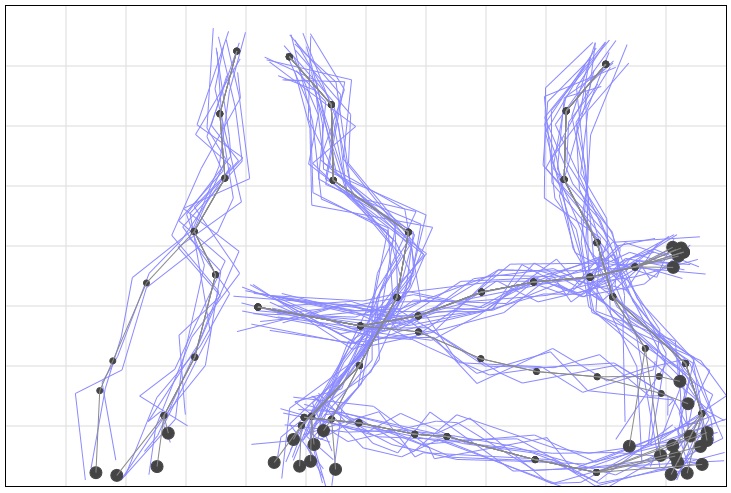}}
  } 
    \caption{ Path diversification methods: (a) square, (b) triangle,
      and (c) triangle.  Original path is shown in Black, and the Blue
      is an example of a diversified path. (d), (e) and (f) shows the
      effects of path diversification.  }
  \label{fig:div_all}
\end{figure}

Crowd analysis returns dominant paths.  If RVO2 is instructed to
simulate agents that move along these paths only, the crowd simulation
will exhibit an ant-like behavior: agents moving along invisible lines
in the scene.  This is undesirable.  This issue is resolved through
path diversification (Figure~\ref{fig:div_all}), which is the process
that generates multiple, slightly different versions of a given path.
Given a single path through the scene, path diversification allows us
to assign a unique path to each agent. Experiments have been conducted
with three methods for generating variations from a given path;
square, triangle, and circle.

The square method defines a square region around each node and
randomly selects a point from the area as the next node (Figure
\ref{fig:div_all} (a)).  This occurs for each node until a diversified
path is generated.  The size of each square needs to selected
carefully to avoid overlap between squares at two adjacent nodes,
which may result in an awkward looking path for each agent.

In triangle method (Figure~\ref{fig:div_all} (b)) the user defines the
size of the base of a triangle drawn at the next node from the current
node.  The base is drawn as the average of perpendicular vectors of
the current and next path segments.  This method is less likely to
suffer from overlap; however, overlap can still occur in some scenarios.

In circle method (Figure~\ref{fig:div_all} (c)) the user defines a
maximum circle size and the algorithm randomly generates circles of
max size or less along the path to the goal.  Sequential circles have
related radii so the resulting paths do not have huge variations.
This method has no overlap.  It should be noted that the resultant
path is randomly placed on one side of the crude path only.  This
logic was added to prevent the generated path from crossing the crude
path and then crossing back.  Criss-crossing can result in a jittery
motion path, which is undesireable.  Radii of adjacent circles are
tied to each other to avoid having the situation where a circle with a
very small radius is sitting next to one with a large radius.  This
prevents overlap and paths that seem to backtrack and loop over
themselves.  The circle method was found to work best and was used as
the method of diversification in this system.

Motion vectors extracted from the exemplar video live in the image
space.  Similarly global paths also live in the 2D image space.  RVO2
agents also live in the 2D space.  This framework aims to synthesize
3D crowds.  This is achieved by making a ground plane assumption:
virtual humans only walk on a (2D) ground plane.  This can be easily
accompanied by back-projecting the global paths (and their variations
generated through path diversification) onto the ground plane.
Back-projection is easy if the location and orientation of the camera
is known with respect to the ground.  This information is sometimes
available for an exemplar video.  In case this information is not
available, the most likely location and orientation of the camera is
selected by observing the exemplar video. This ground plane assumption
has an obvious limitation.  The framework currently only handles
crowds that move in a single plane.

The simulation is rendered into a video that has the same framerate
and resolution as the exemplar video.  Furthermore, for similarity
computations the location and orientation of the camera used to record
synthetic video should be as close to that of the camera used to
record the exemplar video.  We do not assume calibrated cameras;
however, if camera calibration information is available, it can be be
used during similarity computations.

\section{Evaluation and Results}

\subsection{Evaluation}

\begin{figure}
  \centerline{
  \includegraphics[width=1.5in]{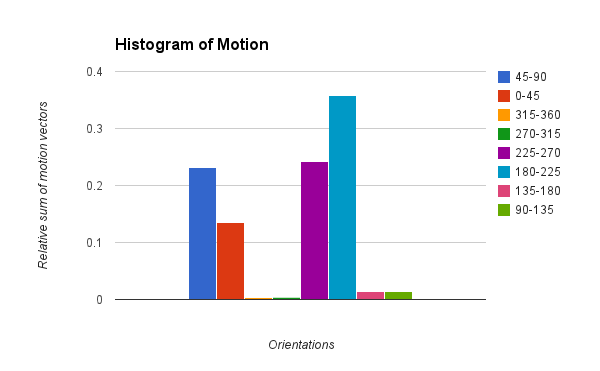}\hspace{.1cm}
  \includegraphics[width=1.5in]{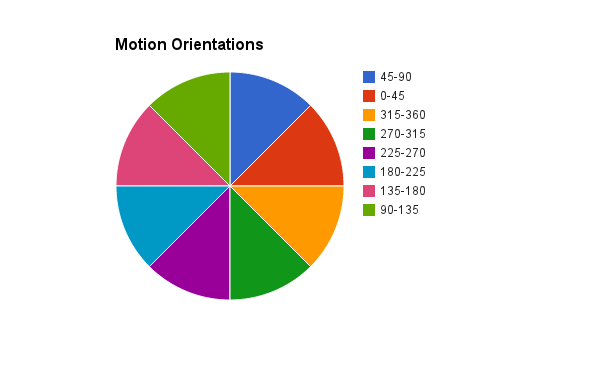}}
  \caption{ (Left) shows a histogram of motion for 8 motion vector orientation ranges.
  (Right) shows the motion vector orientation ranges}
  \label{fig:hom}
\end{figure}

\begin{figure}
  \centerline{
  \includegraphics[width=1.5in]{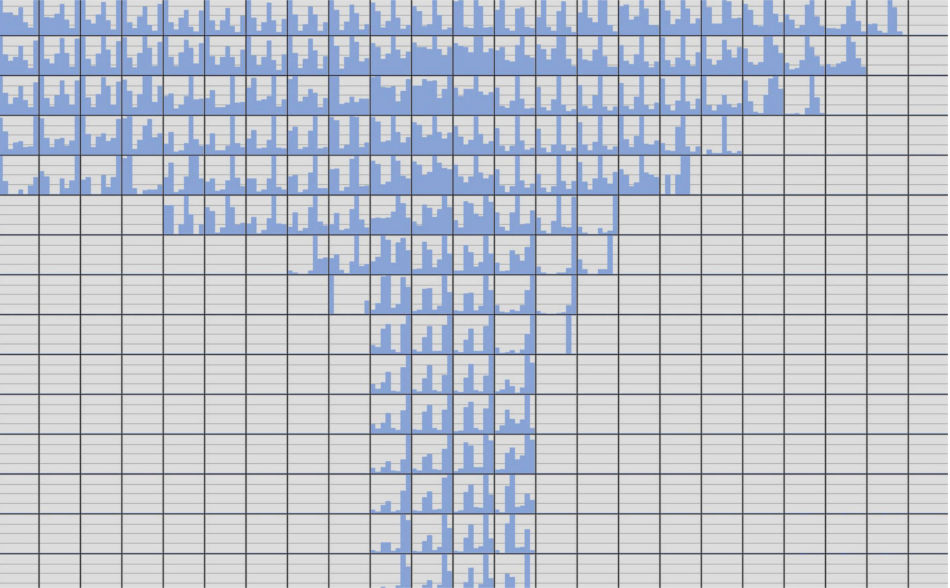}\hfill
  \includegraphics[width=1.5in]{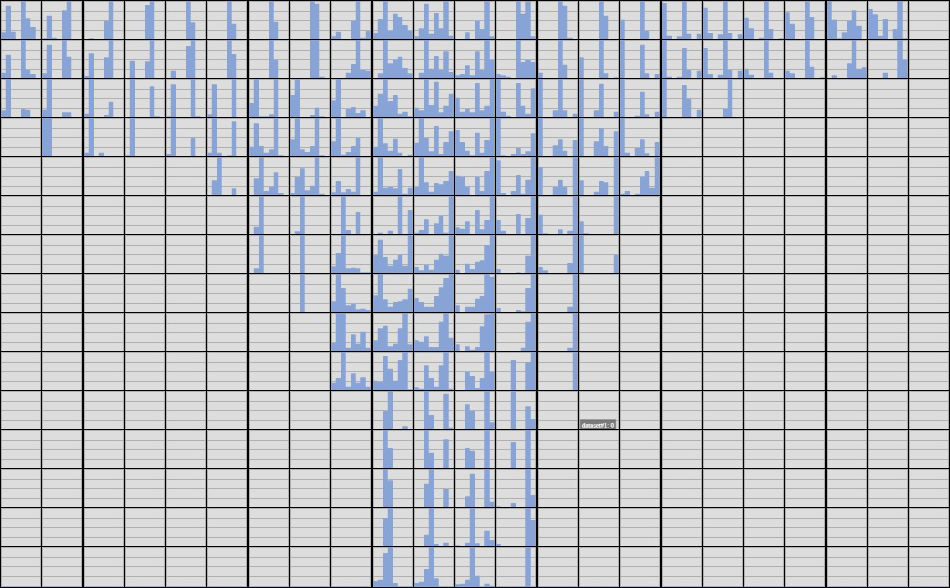}}
  \caption{ Visualization output as a result of the histograms of motion
  generation.  For this scene, histograms of motion were created using 
  a sliding window of size 60x60 pixels.  The window advances 30
  pixels in
  each iteration (50\% overlap). (Left) real Campus dataset video (Right) synthetic 
  result}
  \label{fig:homs}
\end{figure}

To ensure that the synthesized crowd (output) is similar to the input
crowd, a comparison is performed.  One straightforward scheme to
compare the synthesized crowd with the crowd viewed in the exemplar
video is to employ user studies.  That, however, defeats the purpose
of this work--- automated methods for synthesizing crowds from
exemplar videos.  Ideally this system will be able to replace user
studies with a scoring system that can leverage image and video
analysis for crowd comparison.  This will allow for iterative,
self-tuning methods for crowd synthesis in the future.

The scoring system is formed using histograms of motion.  Histograms
of motion show the distribution of motion directions for a given
region in the image (Figure \ref{fig:hom}).  In this system, the
histograms of motion show the distribution of motion flow vectors
using eight orientations.  Recall that the first stage of the
framework is to extract motion vectors.  These collections of motion
flow vectors are what are being compared using the histograms of
motions.

The proposed method subdivides the image into rectangular regions and
generates a series of histograms of motion directions.  This is
similar to how local bins are used to cluster flow vectors based on
orientation and spatial location in the dominant path generation
process.  However, to remove discrete barriers between one histogram
and the next, a sliding window is used to generate the series of
histograms of motion.  These histograms are normalized.  For the
remainder of this section, assume that the sliding window operation
creates $m$ histograms (i.e., unique sub-windows) for each video.

The system outputs a visualization of the histograms of motion (Figure
\ref{fig:homs}).  This visualization makes it easy to spot differences
between scenes and how agents move through them.  This visualization
is good for us, but generating a relative score would prove even more
useful for comparing histograms.  The histograms between real video
data and the synthesized crowd video data are compared using the
Bhattacharyya distance.  The Bhattacharyya distance measures the
dissimilarity between two distributions (histograms), outputting a
value between 0.0 and 1.0 with 0.0 meaning the histograms are a
perfect match and 1.0 meaning the histograms are opposite.  The
Bhattacharyya distance is defined as:
\[
d(H_1,H_2) =  \sum^n_{i=1} \sqrt{ H_1(i) \times H_2(i) },
\]
where $n$ is the number of bins and $H_1(i)$ and $H_2(i)$ are bin
counts for $i^{\mathrm{th}}$ bins of histograms $H_1$ and $H_2$.
The final similarity score between the two videos is:
\[
s(v_1,v_2) = \frac { \sum^m_{i=1} d(H^i_1,H^i_2) }{m},
\] 
where $m$ is the number of histograms extracted for each video (through
sliding window procedure), $H^i_1$ and $H^i_2$ are the $i^{\mathrm{th}}$ histogram
for videos $v_1$  and $v_2$, respectively. 

\subsection{Results}

The proposed framework is tested on 3 crowd videos, each with its own
challenges and intricacies. One video comes from the BIWI Walking
Pedestrian Dataset, the second is the Grand Central Station Dataset,
while the final is from the UCF Crowd Dataset.  All three videos are
recorded from an overhead camera.

Each video is tested in twelve scenarios to see how the proposed
metric performs.  The crowds are tested with three different densities
which are scene specific.  Furthermore the crowds are tested with
tight or loose goals.  This is pertaining to how easily a goal is
achieved.  In the case of tight goals, an agent must be very close to
the goal before they can advance, whereas loose goals are a bit more
forgiving.  This was included as it generally causes problems with the
natural flow of the crowd if goals are too tight.  Agents tend to
circle a goal (i.e., location) or twitch if they are close to a goal and many agents
are near.  This scenario would easily be picked up by a human observer
and labelled as unnatural.  Lastly, the crowds are tested with and
without path diversification.  This test is performed because a human
observer would be able to see agents forming single file lines.  This
tests to see if the metric likes diversified paths or not.

\subsubsection{Campus Video}

\begin{table}[h]\footnotesize
\centering
\begin{tabular}{ | l | l |}
\hline
Synthetic Crowd Characteristics & Score \\ \hline
05 agents, random paths & 0.5506 \\ \hline
10 agents, random paths & 0.5080 \\ \hline
15 agents, random paths & 0.5033 \\ \hline
05 agents, tight goals, no diversification & 0.5462 \\ \hline
10 agents, tight goals, no diversification & 0.4970 \\ \hline
15 agents, tight goals, no diversification & 0.5001 \\ \hline
05 agents, loose goals, no diversification & 0.5487 \\ \hline
10 agents, loose goals, no diversification & 0.5120 \\ \hline
15 agents, loose goals, no diversification & 0.4619 \\ \hline
05 agents, loose goals, diversification & 0.5372 \\ \hline
10 agents, loose goals, diversification & 0.5210 \\ \hline
15 agents, loose goals, diversification & 0.4901 \\
\hline
\end{tabular}
\caption{ Twelve synthetic crowds of the Campus Dataset and their scores based
on Histograms of Motion.}
\label{table:campus}
\end{table}

\begin{figure}
  \centering
  \includegraphics[width=.33\linewidth]{campus_homs_real}\hfill
  \includegraphics[width=.33\linewidth]{campus_homs_best}\hfill
  \includegraphics[width=.33\linewidth]{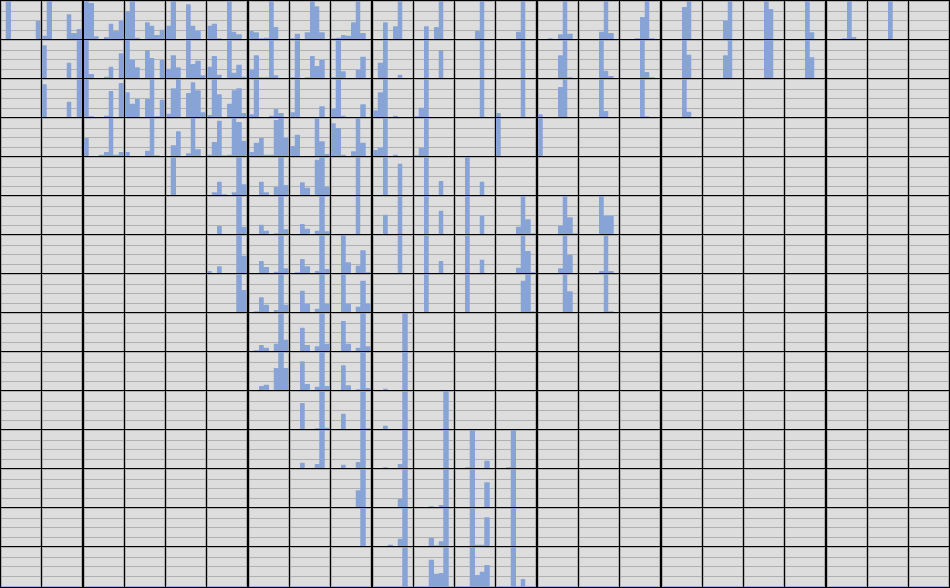}\\
  \caption{ Campus video histograms. (Left) Histograms of Motion from the real video. (Middle) Histograms
  of motion from the best (15 agents, loose goals, no diversification) synthetic crowd.  (Right) Histograms
  of motion from the worst (05 agents, random paths) synthetic crowd.}
  \label{fig:campus_homs}
\end{figure}

A population size of 10 agents was arbitrarily chosen as a starting
point from the observed Campus video, 05 and 15 are values selected
relative to this.  The results are shown in Table \ref{table:campus}.
The only characteristic that can be seen to consistently beat the
random case on this dataset is when using 15 pedestrians.  The 15
pedestrian scenarios seem to produce the best scores, and best score
overall, 0.4619, was accomplished with 15 agents operating with loose
goals and no path diversification.  The worst score was found to be 05
agents operating on random paths, the score being 0.5506.

\subsubsection{Grand Central Video}

\begin{table}[h]\footnotesize
\centering
\begin{tabular}{ | l | l |}
\hline
Synthetic Crowd Characteristics & Score \\ \hline
50 agents, random paths & 0.4539 \\ \hline
75 agents, random paths & 0.5048 \\ \hline
100 agents, random paths & 0.4421 \\ \hline
50 agents, tight goals, no diversification & 0.6091 \\ \hline
75 agents, tight goals, no diversification & 0.5790 \\ \hline
100 agents, tight goals, no diversification & 0.5677 \\ \hline
50 agents, loose goals, no diversification & 0.5844 \\ \hline
75 agents, loose goals, no diversification & 0.5845 \\ \hline
100 agents, loose goals, no diversification & 0.5808 \\ \hline
50 agents, loose goals, diversification & 0.4866 \\ \hline
75 agents, loose goals, diversification & 0.5624 \\ \hline
100 agents, loose goals, diversification & 0.4082 \\
\hline
\end{tabular}
\caption{ Twelve synthetic crowds of the Grand Central Dataset and their scores based
on Histograms of Motion.}
\label{table:grand}
\end{table}

\begin{figure}
  \centering
  \includegraphics[width=.33\linewidth]{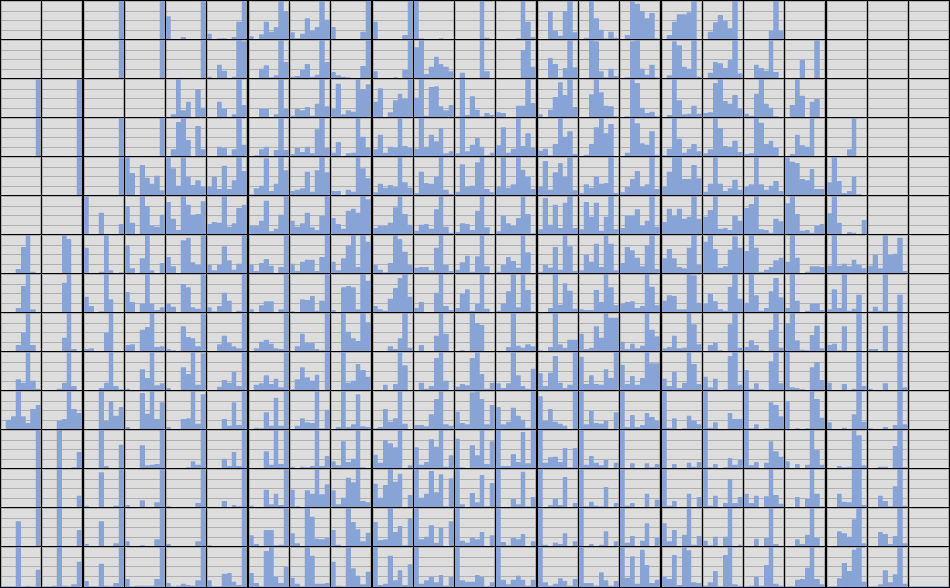}\hfill
  \includegraphics[width=.33\linewidth]{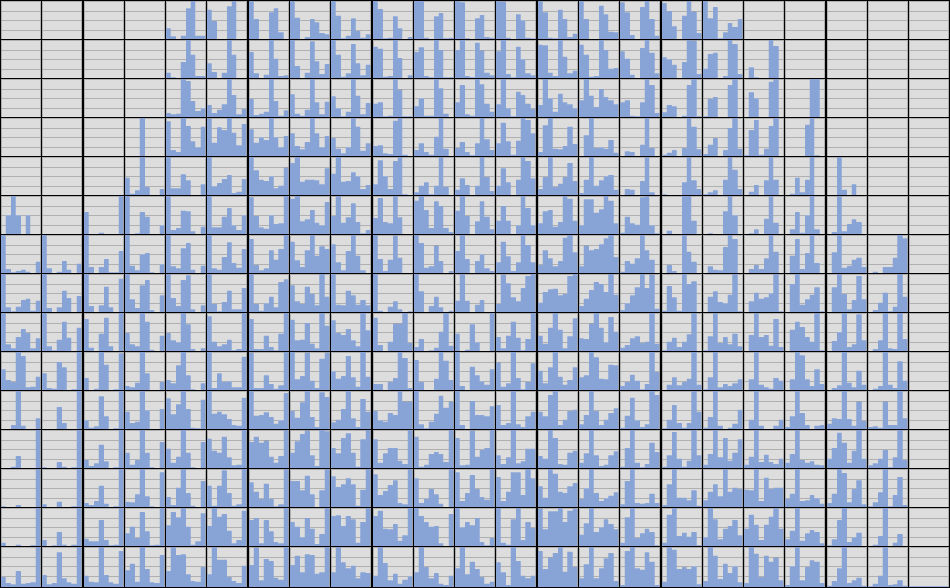}\hfill
  \includegraphics[width=.33\linewidth]{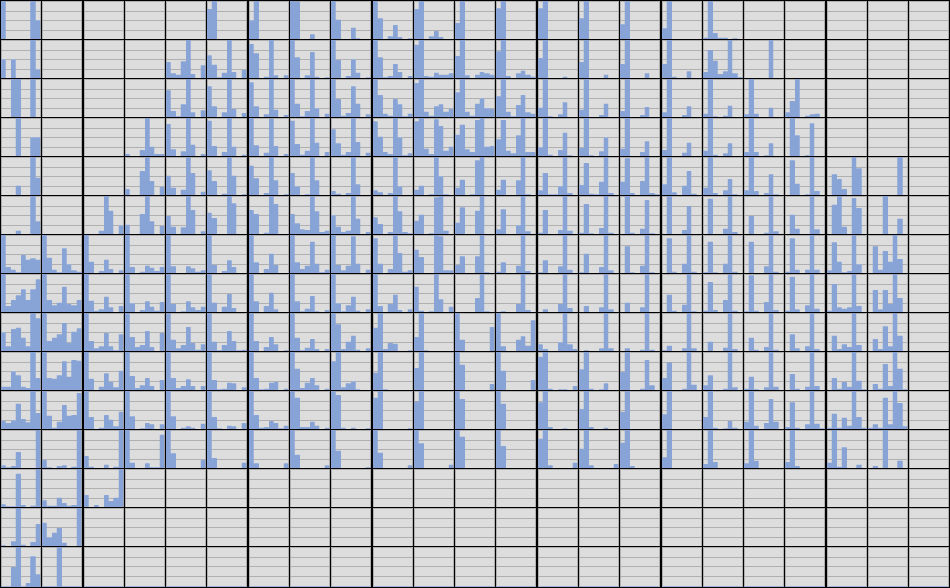}\\
  \caption{ Grand central video histograms. (Left) Histograms of Motion from the real video. (Middle) Histograms
  of motion from the best (15 agents, loose goals, no diversification) synthetic crowd.  (Right) Histograms
  of motion from the worst (05 agents, random paths) synthetic crowd.}
  \label{fig:campus_homs}
\end{figure}

The crowd population sizes (50,75,100) in these tests are much larger
than the other two data sets.  The results can be seen in Table
\ref{table:grand}.  This dataset has mixed results.  The worst
performing scenario being 50 agents operating with tight goals and no
diversification.  The best performing scenario being 100 agents
operating with loose goals and diversification.  The worst and best
scores being 0.6091 and 0.4082 respectively.  The best score obtained
with this dataset is what should be expected from human trials as the
agents flow from point to point more naturally while still following
the dominant path logic observed in the input video.  This is
promising.  However, the expected worst case would be random paths.
Upon visual inspection, we found that the input video did exhibit random crowd motion.

\subsubsection{UCF Crowd Video}

\begin{table}[h]\footnotesize
\centering
\begin{tabular}{ | l | l |}
\hline
Synthetic Crowd Characteristics & Score \\ \hline
10 agents, random paths & 0.2573 \\ \hline
20 agents, random paths & 0.2428 \\ \hline
30 agents, random paths & 0.2423 \\ \hline
10 agents, tight goals, no diversification & 0.3774 \\ \hline
20 agents, tight goals, no diversification & 0.2858 \\ \hline
30 agents, tight goals, no diversification & 0.4390 \\ \hline
10 agents, loose goals, no diversification & 0.3508 \\ \hline
20 agents, loose goals, no diversification & 0.3328 \\ \hline
30 agents, loose goals, no diversification & 0.2436 \\ \hline
10 agents, loose goals, diversification & 0.3254 \\ \hline
20 agents, loose goals, diversification & 0.2949 \\ \hline
30 agents, loose goals, diversification & 0.3026 \\
\hline
\end{tabular}
\caption{ Twelve synthetic crowds of the UCF Crowd Dataset and their scores based
on Histograms of Motion.}
\label{table:ucf}
\end{table}

\begin{figure}
  \centering
  \includegraphics[width=.33\linewidth]{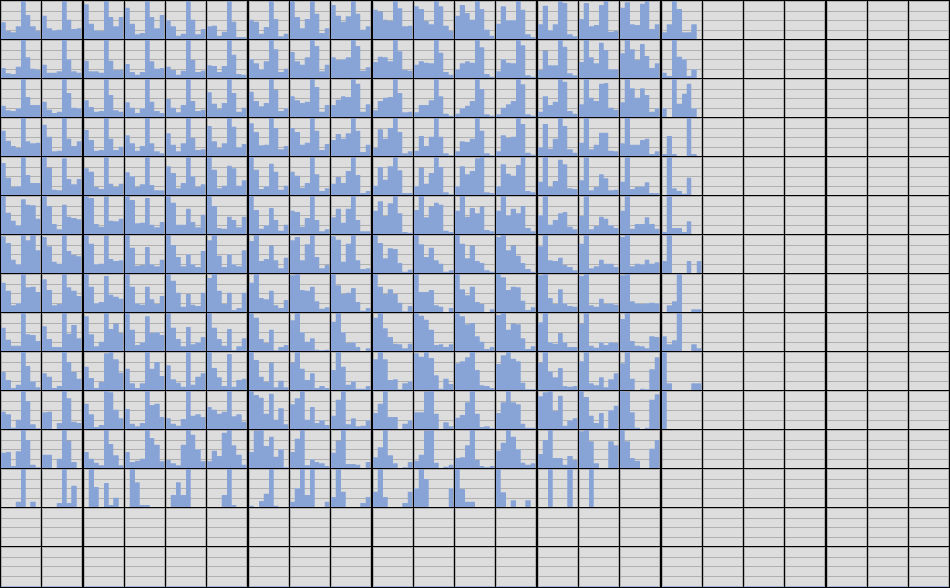}\hfill
  \includegraphics[width=.33\linewidth]{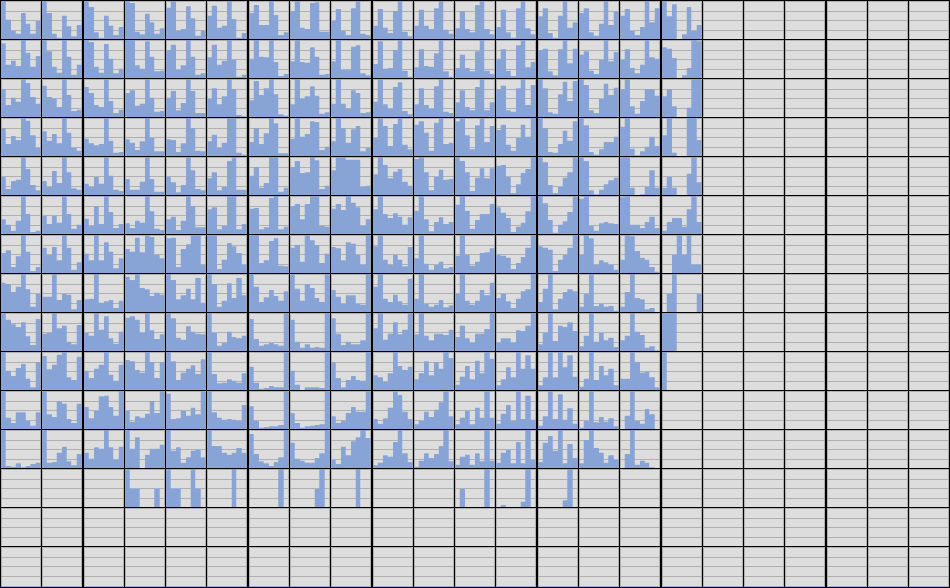}\hfill
  \includegraphics[width=.33\linewidth]{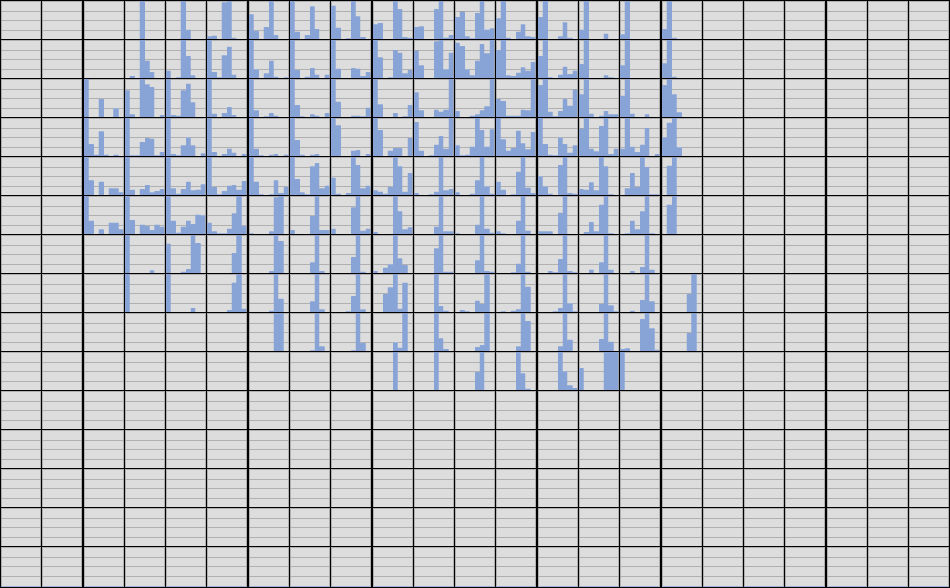}\\
  \caption{ UCF video histograms. (Left) Histograms of Motion from the real video. (Middle) Histograms
  of motion from the best (30 agents, random paths) synthetic crowd.  (Right) Histograms
  of motion from the worst (30 agents, tight goals, no diversification) synthetic crowd.}
  \label{fig:campus_homs}
\end{figure}

Results (Table \ref{table:ucf}) are mixed, and the metric generally
favours random motions.  The metric performs poorly on this dataset.
However, the original video has very unpredictable motions for
pedestrians.  The best performing synthetic scenario is 30 agents
operating on random paths, resulting in a score of 0.2423.  The worst
performing scenario was 30 agents operating with tight goals and no
path diversification.  The difference between the best and worst score
makes sense as the input video has a very random naturally occurring
crowd while the tight goals and no diversification simulation is very
restricted.  However, in this scenario random paths performed the
best.  It should be noted that the difference in score between the
best random scenario and the best processed scenario is only 0.0013.

\subsection{Interacting with Synthesized Crowds} 

A strength of this framework is the ease at which the environment can
be manipulated.  Items (obstacles) can be moved, added, or removed,
and the synthesized crowd responds appropriately.  Similarly camera
angles can be changed to alter how a crowd is observed.  These
manipulations can be used to test changes in simulated scenarios.  For
example, testing evacuation scenarios with different obstacles
involved.  We leave this for aspect of our work for another time.
Figure~\ref{fig:homs} shows how crowd responds to changes in its environment.

\begin{figure*}
  \centerline{
    \includegraphics[width=.25\linewidth]{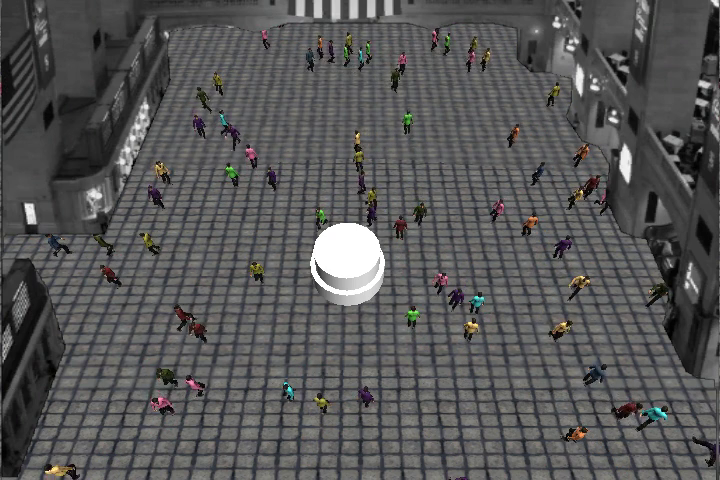}\hfill
  \includegraphics[width=.25\linewidth]{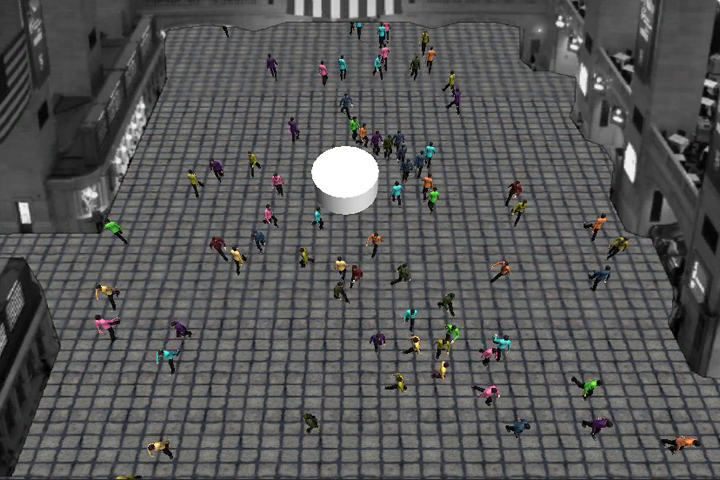}\hfill
  \includegraphics[width=.25\linewidth]{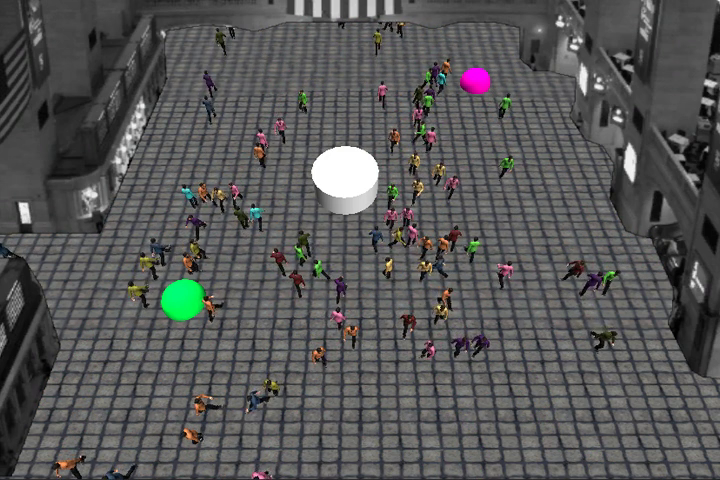}\hfill
  \includegraphics[width=.25\linewidth]{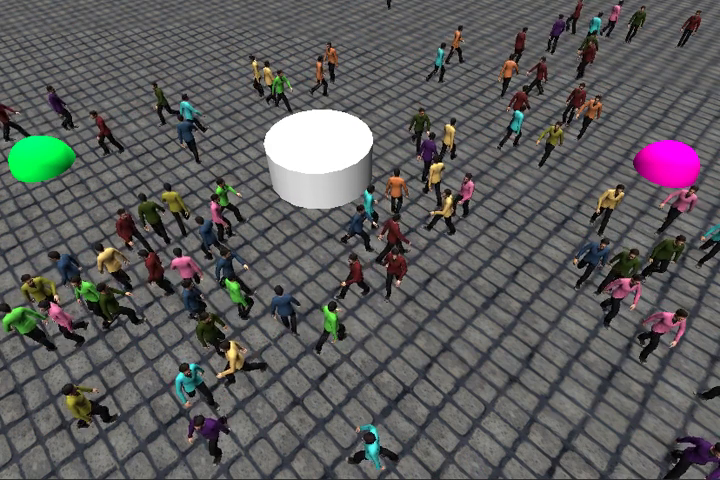}
  }
  \caption{ Synthesized crowd responds to changes in its
    environment.  (Left to right) original setting, the central
    obstacle has been moved to a new location, and two new obstacles added
    to the scene. (right most) crowd seen from a different viewpoint.}
  \label{fig:manip}
\end{figure*}


\section{Conclusions and Future Work}

This paper presents a vision-based crowd synthesis framework.  Motion
extracted from crowd videos is used to control a set of 3D virtual
pedestrians.  Each virtual pedestrian is a self-animating autonomous
agent.  Motion histograms extracted from input crowd video is compared
with those extracted from synthesized crowd video to compute a
similarty score, which captures how faithfully synthesized crowd
exhibits the motion characteristics of the crowd seen in the input
video.  In the future we plan to use this similarity score to
auto-tune crowd synthesis, i.e., tweaking behavior parameters to
improve the fidelty of the synthesized crowds with respect to one or
more input videos.  In order to achieve this end, we will need
mechanisms for automatic camera pose and ground plane detection.

Using the histogram of motion has mixed results as a metric for
determining which crowd is the best synthesized version of a real
crowd.  From preliminary results presented here, it seems that the
metric provides a mechanism to rank videos according to their
similarity to the exemplar videos.  There are some limitations to the
metric.  Firstly, temporal information is disregarded in the global
motion vector collection process which is used for comparison between
scenes.  This could present a problem in time sensitive scenarios such
as a cross walk at a red light.  We plan to address this limitation in the future.

The current set of animations is very basic.  Agents are able to
transition between walk and run states, with turns either left or
right.  They are also able to perform an idle animation whereby they
shift their weight and perform subtle motions while remaining in a
single position.  Having a more intricate set of animations is
desirable for this system.  Incorporating some form of interaction
between agents (talk gestures, shake hands, etc.) would help the
simulation feel more realistic.  Similarly having agents interact with
the environment (sit on bench, stop and look at something in the
scene, react to emergency) would contribute to the feeling of a
reality and further improve the fidelity of the animations.

\bibliographystyle{IEEEtran}
\bibliography{crowds.bib}

\begin{thebibliography}{10}
\providecommand{\url}[1]{#1}
\csname url@samestyle\endcsname
\providecommand{\newblock}{\relax}
\providecommand{\bibinfo}[2]{#2}
\providecommand{\BIBentrySTDinterwordspacing}{\spaceskip=0pt\relax}
\providecommand{\BIBentryALTinterwordstretchfactor}{4}
\providecommand{\BIBentryALTinterwordspacing}{\spaceskip=\fontdimen2\font plus
\BIBentryALTinterwordstretchfactor\fontdimen3\font minus
  \fontdimen4\font\relax}
\providecommand{\BIBforeignlanguage}[2]{{%
\expandafter\ifx\csname l@#1\endcsname\relax
\typeout{** WARNING: IEEEtran.bst: No hyphenation pattern has been}%
\typeout{** loaded for the language `#1'. Using the pattern for}%
\typeout{** the default language instead.}%
\else
\language=\csname l@#1\endcsname
\fi
#2}}
\providecommand{\BIBdecl}{\relax}
\BIBdecl

\bibitem{reynolds1987flocks}
C.~W. Reynolds, ``Flocks, herds and schools: A distributed behavioral model,''
  \emph{ACM SIGGRAPH Computer Graphics}, vol.~21, pp. 25--34, 1987.

\bibitem{massive14}
\BIBentryALTinterwordspacing
``Massive software - simulating life,'' Aug. 2014. [Online]. Available:
  \url{http://www.massivesoftware.com/}
\BIBentrySTDinterwordspacing

\bibitem{golaem14}
\BIBentryALTinterwordspacing
``Crowd simulation for maya | golaem crowd,'' Aug. 2014. [Online]. Available:
  \url{http://www.golaem.com/}
\BIBentrySTDinterwordspacing

\bibitem{miarmy14}
\BIBentryALTinterwordspacing
``Miarmy - basefount miarmy maya crowd simulation tools,'' Aug. 2014. [Online].
  Available: \url{http://www.basefount.com/miarmy/}
\BIBentrySTDinterwordspacing

\bibitem{maya14}
\BIBentryALTinterwordspacing
``3d animation software, computer animation software | maya | autodesk,'' Aug.
  2014. [Online]. Available:
  \url{http://www.autodesk.com/products/maya/overview}
\BIBentrySTDinterwordspacing

\bibitem{li2012cloning}
Y.~Li, M.~Christie, O.~Siret, R.~Kulpa, and J.~Pettr{\'e}, ``Cloning crowd
  motions,'' in \emph{Proceedings of the ACM SIGGRAPH/Eurographics Symposium on
  Computer Animation}.\hskip 1em plus 0.5em minus 0.4em\relax Eurographics
  Association, 2012, pp. 201--210.

\bibitem{van2008reciprocal}
J.~Van~den Berg, M.~Lin, and D.~Manocha, ``Reciprocal velocity obstacles for
  real-time multi-agent navigation,'' in \emph{Robotics and Automation, 2008.
  ICRA 2008. IEEE International Conference on}.\hskip 1em plus 0.5em minus
  0.4em\relax IEEE, 2008, pp. 1928--1935.

\bibitem{kovar2002motion}
L.~Kovar, M.~Gleicher, and F.~Pighin, ``Motion graphs,'' \emph{ACM transactions
  on graphics (TOG)}, vol.~21, pp. 473--482, 2002.

\bibitem{sumpter2006principles}
D.~J. Sumpter, ``The principles of collective animal behaviour,''
  \emph{Philosophical Transactions of the Royal Society B: Biological
  Sciences}, vol. 361, pp. 5--22, 2006.

\bibitem{bode2010perceived}
N.~W. Bode, J.~J. Faria, D.~W. Franks, J.~Krause, and A.~J. Wood, ``How
  perceived threat increases synchronization in collectively moving animal
  groups,'' \emph{Proceedings of the Royal Society B: Biological Sciences},
  vol. 277, pp. 3065--3070, 2010.

\bibitem{thornton2004incidental}
I.~M. Thornton and Q.~C. Vuong, ``Incidental processing of biological motion,''
  \emph{Current Biology}, vol.~14, pp. 1084--1089, 2004.

\bibitem{eshel2010tracking}
R.~Eshel and Y.~Moses, ``Tracking in a dense crowd using multiple cameras,''
  \emph{International journal of computer vision}, vol.~88, no.~1, pp.
  129--143, 2010.

\bibitem{hu2008learning}
M.~Hu, S.~Ali, and M.~Shah, ``Learning motion patterns in crowded scenes using
  motion flow field,'' in \emph{ICPR}, 2008, pp. 1--5.

\bibitem{ozturk2010detecting}
O.~Ozturk, T.~Yamasaki, and K.~Aizawa, ``Detecting dominant motion flows in
  unstructured/structured crowd scenes,'' in \emph{Pattern Recognition (ICPR),
  2010 20th International Conference on}.\hskip 1em plus 0.5em minus
  0.4em\relax IEEE, 2010, pp. 3533--3536.

\bibitem{yersin2009crowd}
B.~Yersin, J.~Ma{\"\i}m, J.~Pettr{\'e}, and D.~Thalmann, ``Crowd patches:
  populating large-scale virtual environments for real-time applications,'' in
  \emph{Proceedings of the 2009 symposium on Interactive 3D graphics and
  games}.\hskip 1em plus 0.5em minus 0.4em\relax ACM, 2009, pp. 207--214.

\bibitem{shum2008interaction}
H.~P. Shum, T.~Komura, M.~Shiraishi, and S.~Yamazaki, ``Interaction patches for
  multi-character animation,'' in \emph{ACM Transactions on Graphics (TOG)},
  vol.~27.\hskip 1em plus 0.5em minus 0.4em\relax ACM, 2008, p. 114.

\bibitem{kim2012tiling}
M.~Kim, Y.~Hwang, K.~Hyun, and J.~Lee, ``Tiling motion patches,'' in
  \emph{Proceedings of the 11th ACM SIGGRAPH/Eurographics conference on
  Computer Animation}.\hskip 1em plus 0.5em minus 0.4em\relax Eurographics
  Association, 2012, pp. 117--126.

\bibitem{lee2006motion}
K.~H. Lee, M.~G. Choi, and J.~Lee, ``Motion patches: building blocks for
  virtual environments annotated with motion data,'' in \emph{ACM Transactions
  on Graphics (TOG)}, vol.~25.\hskip 1em plus 0.5em minus 0.4em\relax ACM,
  2006, pp. 898--906.

\bibitem{patil2011directing}
S.~Patil, J.~Van Den~Berg, S.~Curtis, M.~C. Lin, and D.~Manocha, ``Directing
  crowd simulations using navigation fields,'' \emph{Visualization and Computer
  Graphics, IEEE Transactions on}, vol.~17, pp. 244--254, 2011.

\bibitem{wang2008data}
X.~Wang and S.~Sun, ``Data-driven macroscopic crowd animation synthesis method
  using velocity fields,'' in \emph{Computational Intelligence and Design,
  2008. ISCID'08. International Symposium on}, vol.~2.\hskip 1em plus 0.5em
  minus 0.4em\relax IEEE, 2008, pp. 157--160.

\bibitem{chenney2004flow}
S.~Chenney, ``Flow tiles,'' in \emph{Proceedings of the 2004 ACM
  SIGGRAPH/Eurographics symposium on Computer animation}.\hskip 1em plus 0.5em
  minus 0.4em\relax Eurographics Association, 2004, pp. 233--242.

\bibitem{flagg2013video}
M.~Flagg and J.~M. Rehg, ``Video-based crowd synthesis,'' \emph{Visualization
  and Computer Graphics, IEEE Transactions on}, vol.~19, no.~11, pp.
  1935--1947, 2013.

\bibitem{butenuth2011integrating}
M.~Butenuth, F.~Burkert, F.~Schmidt, S.~Hinz, D.~Hartmann, A.~Kneidl,
  A.~Borrmann, and B.~Sirma{\c{c}}ek, ``Integrating pedestrian simulation,
  tracking and event detection for crowd analysis,'' in \emph{Computer Vision
  Workshops (ICCV Workshops), 2011 IEEE International Conference on}.\hskip 1em
  plus 0.5em minus 0.4em\relax IEEE, 2011, pp. 150--157.

\bibitem{lee2007group}
K.~H. Lee, M.~G. Choi, Q.~Hong, and J.~Lee, ``Group behavior from video: a
  data-driven approach to crowd simulation,'' in \emph{Proceedings of the 2007
  ACM SIGGRAPH/Eurographics symposium on Computer animation}.\hskip 1em plus
  0.5em minus 0.4em\relax Eurographics Association, 2007, pp. 109--118.

\bibitem{lerner2009fitting}
A.~Lerner, E.~Fitusi, Y.~Chrysanthou, and D.~Cohen-Or, ``Fitting behaviors to
  pedestrian simulations,'' in \emph{Proceedings of the 2009 ACM
  SIGGRAPH/Eurographics Symposium on Computer Animation}.\hskip 1em plus 0.5em
  minus 0.4em\relax ACM, 2009, pp. 199--208.

\bibitem{sun2011data}
L.~Sun and W.~Qin, ``A data-driven approach for simulating pedestrian collision
  avoidance in crossroads,'' in \emph{Digital Media and Digital Content
  Management (DMDCM), 2011 Workshop on}.\hskip 1em plus 0.5em minus 0.4em\relax
  IEEE, 2011, pp. 83--85.

\bibitem{guy2012statistical}
S.~J. Guy, J.~van~den Berg, W.~Liu, R.~Lau, M.~C. Lin, and D.~Manocha, ``A
  statistical similarity measure for aggregate crowd dynamics,'' \emph{ACM
  Transactions on Graphics (TOG)}, vol.~31, no.~6, p. 190, 2012.

\bibitem{ihaddadene2008real}
N.~Ihaddadene and C.~Djeraba, ``Real-time crowd motion analysis,'' in
  \emph{Pattern Recognition, 2008. ICPR 2008. 19th International Conference
  on}.\hskip 1em plus 0.5em minus 0.4em\relax IEEE, 2008, pp. 1--4.

\bibitem{Lee:2007:GBV:1272690.1272706}
\BIBentryALTinterwordspacing
K.~H. Lee, M.~G. Choi, Q.~Hong, and J.~Lee, ``Group behavior from video: A
  data-driven approach to crowd simulation,'' in \emph{Proceedings of the 2007
  ACM SIGGRAPH/Eurographics Symposium on Computer Animation}, ser. SCA
  '07.\hskip 1em plus 0.5em minus 0.4em\relax Aire-la-Ville, Switzerland,
  Switzerland: Eurographics Association, 2007, pp. 109--118. [Online].
  Available: \url{http://dl.acm.org/citation.cfm?id=1272690.1272706}
\BIBentrySTDinterwordspacing

\bibitem{karamouzas2012simulating}
I.~Karamouzas and M.~Overmars, ``Simulating and evaluating the local behavior
  of small pedestrian groups,'' \emph{Visualization and Computer Graphics, IEEE
  Transactions on}, vol.~18, no.~3, pp. 394--406, 2012.

\bibitem{farneback2003two}
G.~Farneb{\"a}ck, ``Two-frame motion estimation based on polynomial
  expansion,'' in \emph{Image Analysis}.\hskip 1em plus 0.5em minus 0.4em\relax
  Springer, 2003, pp. 363--370.

\bibitem{zelnik2004self}
L.~Zelnik-Manor and P.~Perona, ``Self-tuning spectral clustering.'' in
  \emph{NIPS}, vol.~17, 2004, p.~16.

\end{thebibliography}

\end{document}